\documentclass[letterpaper]{article} 
\usepackage{aaai2026}  
\usepackage{times}  
\usepackage{helvet}  
\usepackage{courier}  
\usepackage[hyphens]{url}  
\usepackage{graphicx} 
\usepackage{natbib}  
\usepackage{caption} 
\usepackage{algorithm}
\usepackage{algorithmic}
\usepackage{enumitem}
\usepackage{amsmath}
\usepackage{amssymb}
\usepackage{booktabs}
\usepackage{siunitx}
\usepackage{makecell}   
\usepackage{booktabs,tabularx,siunitx,array}
\newcolumntype{Y}{>{\centering\arraybackslash}p{1.2em}} 
\usepackage{placeins}

\usepackage{array}    
\usepackage{amssymb}  
\newcolumntype{L}[1]{>{\raggedright\arraybackslash}p{#1}}

\usepackage{newfloat}
\usepackage{listings}
\DeclareCaptionStyle{ruled}{labelfont=normalfont,labelsep=colon,strut=off} 
\floatstyle{ruled}
\newfloat{listing}{tb}{lst}{}
\floatname{listing}{Listing}
\title{LLM-Based Agents for Competitive Landscape Mapping \\ in Drug Asset Due Diligence}

\author{
    Vlad Vinogradov\textsuperscript{\rm 1},
    Alisa Vinogradova\textsuperscript{\rm 3},
    Dmitrii Radkevich\textsuperscript{\rm 1},
    Ilya Yasny\textsuperscript{\rm 1, \rm 2},
    Dmitry Kobyzev\textsuperscript{\rm 1, \rm 2}, \\
    Ivan Izmailov\textsuperscript{\rm 1},
    Katsiaryna Yanchanka\textsuperscript{\rm 3},
    Roman Doronin\textsuperscript{\rm 1},
    Andrey Doronichev\textsuperscript{\rm 1}
}
\affiliations{
    \textsuperscript{\rm 1}Bioptic.io, San Francisco, CA\\
    \textsuperscript{\rm 2}LanceBio Ventures\\
    \textsuperscript{\rm 3}AI Expert\\

    info@optic.inc
}

\usepackage{bibentry}

\begin{document}

\maketitle



\makeatletter
\makeatother

\begin{abstract}
In this paper, we describe and benchmark a competitor-discovery component used within an agentic AI system for fast drug asset due diligence. A competitor-discovery AI agent, given an indication, retrieves all drugs comprising the competitive landscape of that indication and extracts canonical attributes for these drugs. The competitor definition is investor-specific, and data is paywalled/licensed, fragmented across registries, ontology-mismatched by indication, alias-heavy for drug names, multimodal, and rapidly changing. Although considered the best tool for this problem, the current LLM-based AI systems aren’t capable of reliably retrieving all competing drug names, and there is no accepted public benchmark for this task. To address the lack of evaluation, we use LLM-based agents to transform five years of multimodal, unstructured diligence memos from a private biotech VC fund into a structured evaluation corpus mapping indications to competitor drugs with normalized attributes. We also introduce a competitor validating LLM-as-a-judge agent that filters out false positives from the list of predicted competitors to maximize precision and suppress hallucinations. In this benchmark, our competitor discovery agent (\textsc{Bioptic Agent}) achieves 83\% recall, exceeding OpenAI Deep Research (65\%) and Perplexity Labs (60\%). The system is deployed in production with enterprise users; in a case study with a biotech VC investment fund, analyst turnaround time dropped from 2.5 days to $\sim$3 hours ($\sim$20x) for the competitive analysis.
\end{abstract}

\begin{links}
\link{Platform}{https://bioptic.io}
\end{links}


\section{Introduction}

Mapping the entire competitive landscape for a drug is the first and important step of the competitive analysis underlying the patent, business, and scientific due diligence for in-/out-licensing, as well as for identifying gaps when planning a clinical trial for that specific asset. Regulators make competitor discovery increasingly important, since 12 January 2025, the EU Health Technology Assessment Regulation applies Joint Clinical Assessments (JCAs) \cite{EU_JCA_Web_2025} for new oncology medicines and all Advanced Therapy Medicinal Products (ATMPs). Missing a relevant competitor or proposing an inappropriate one can lead the JCA to set comparators that the drug developer did not anticipate, exposing the asset to European Union (EU)-wide market access and pricing risk; previously, such exposure was confined to national decisions. 
Globally, expectations align: ICH E10 \cite{ICH_E10_2000} and FDA effectiveness guidances \cite{FDA_SE_OneStudy_2023} emphasize appropriate control/comparator selection, and NICE \cite{NICE_PMG36_2022} methods require justifiable comparators in routine practice.

\FloatBarrier
\begin{figure}[!t]
  \centering
  \includegraphics[width=1.0\linewidth]{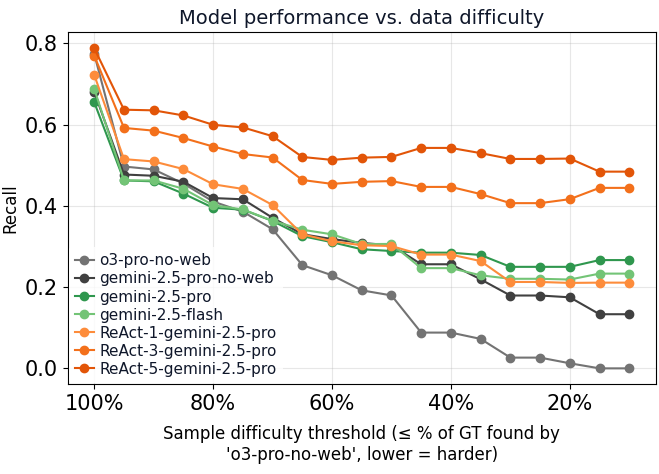}
  \caption{
    Model performance across varying levels of sample difficulty.
    The x-axis denotes difficulty thresholds: at each point, we evaluate all models on the subset of samples where \textit{o3-pro-no-web} recovered $\leq$ the indicated percentage of ground-truth competitors.
    This allows us to assess how different agents perform on increasingly difficult samples, using a non-web baseline as the difficulty proxy.
  }
  \label{fig:sample-difficulty}
\end{figure}

\begin{figure*}[t]
  \centering
  \includegraphics[width=17
cm]{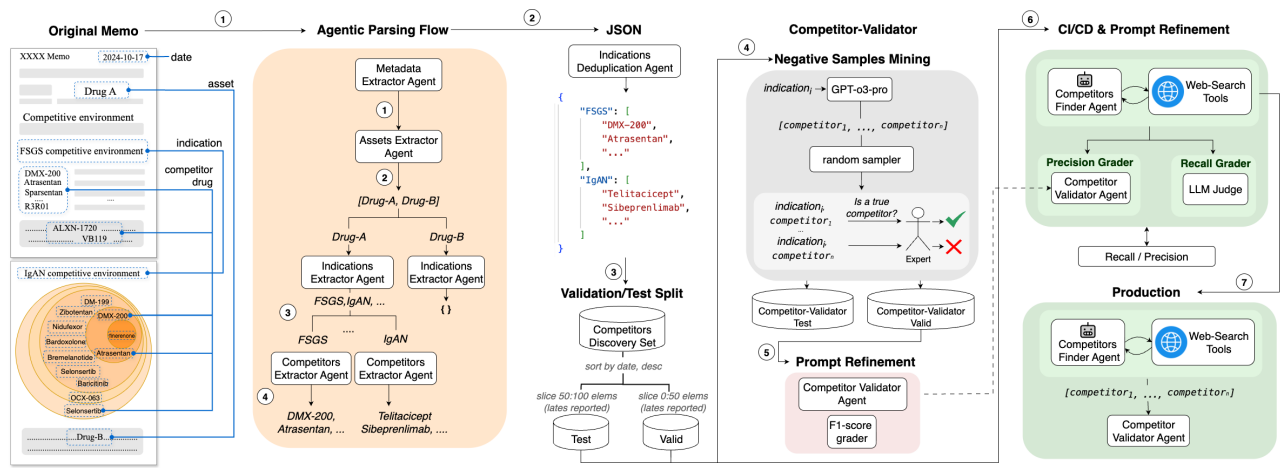}
  \caption{Competitors Discovery System Diagram. Diligence memos are parsed by a hierarchical multi-agent system into normalized JSON (assets, indications, competitors), deduplicated, and split chronologically. Precision is graded by an expert-trained LLM-as-a-judge Competitor-Validator with mined negatives, recall by an LLM-as-a-judge that uses Web-search; both feed CI/CD. In production, a Competitors-Finder agent calls the web tools, and the Validator filters candidates.}
  \label{fig:framework-diagram}
\end{figure*}

Identifying all competitors for an indication demands consolidating evidence from press releases, scientific literature, patents, clinical-trial registries, and target-space intelligence, then linking novel targets to the drugs that modulate them. Because information is scattered, alias-ridden, multilingual, and ranked differently by region, and the landscape changes rapidly without central tracking, teams still rely on manual expert workflows and spreadsheets, creating a persistent completeness, recency, and cycle-time risks that cascade into comparator strategy, trial design, pricing, and licensing decisions.

Large language models (LLMs) appear well-suited to this task: they are increasingly multimodal and multilingual, can automate operating with scientific tools, web browsing, managing external databases at scale, and are being rapidly operationalized across big pharma. For example, Moderna shifted from an internal GPT-4 API agent (“mChat”) to ChatGPT Enterprise for thousands of employees, spawning 750+ internal GPTs including “Dose ID” \cite{Moderna2024OpenAI}; Takeda deployed a secure Azure OpenAI assistant with PwC/Microsoft \cite{PwC2024Takeda1}; and Bayer’s ChatGPT-4 Turbo–based MyGenAssist cut pharmacovigilance letter time by ~23\% \cite{Benaiche2025MyGenAssist}. Moreover, recent studies report that general-purpose, frontier LLMs can be competitive with—or even outperform—task-specific fine-tuned models on certain pharma-adjacent evaluations \cite{Chen2025NatComm_BioNLP, McDuff2025Nature_DxLLMs}. 
Accordingly, since marginal gains in this setting typically come from agentic scaffolds and execution policy rather than weight adjustments to a specific task, and apparent prompt wins can degrade under real-world constraints, it is necessary to prioritize end-to-end, domain-grounded benchmarking that exercises failure modes, to ensure deployed systems remain stable and maintain certain Service Level Agreements (SLAs).

There is no publicly accepted benchmark for competitor discovery because key evidence and curated lists are paywalled and heterogeneous across providers (e.g., Clarivate, Citeline, GlobalData, AlphaSense), necessitating in-house expert curation, and the definition of a “competitor” is investor- and stakeholder-specific. This task heavily relies on web browsing and multi-hop navigation across different sources, backtracking, and evidence reconciliation. 
Web-agent benchmarks such as BrowseComp (hard-to-surface web facts) \cite{wei2025browsecomp}, WebVoyager (interacting with real-world websites) \cite{he2024webvoyagerbuildingendtoendweb}, WebArena (realistic but closed-world sites) \cite{zhou2023webarena}, Mind2Web (generalist website tasks) \cite{deng2023mind2web}, and WebLINX (conversational web navigation) \cite{lu2024weblinx} are not sufficient to capture the complexity of the competitors discovery problem: nomenclature harmonization; landscape completeness, long-tail coverage; and high-recall retrieval at regulator-grade. To fill this gap, we use LLM-based agents to transform five years of multi-modal, unstructured diligence memos from a private biotech VC fund into a structured evaluation corpus that maps indications to competitor drugs with normalized attributes. 

Because the primary deployment risk is incompleteness, our benchmark first makes recall observable — how many expert-identified competitors the agent recovers. At the same time, recent work presenting ChiDrug \cite{wu-etal-2025-believe}, a benchmark comprising six Chinese medication subtasks, including an indication task, shows frontier models (e.g., GPT-4o, Claude 3.5) both omit valid answers (recall failures) and hallucinate (precision failures), underscoring the need to measure completeness and to validate outputs. We therefore introduce the post-retrieval filtering LLM-as-judge (Competitors Validator), which, in the fashion of the PaperBench’s “expert slice” \cite{starace2025paperbench}, uses a conservative synthetic-labeling step in which a calibrated agent, tuned to a 90\% F1-score of detecting false-positive competitor drugs, proposes false candidates that are then filtered out before propagating them to the client-facing components of the system.

Another challenge in productionizing LLM applications is balancing the “bitter lesson” (that durable gains come from general methods that scale rather than hand-crafted domain rules) with the empirical need for agentic scaffolds on long-tail, multi-hop web tasks. As single‑pass prompting, even with comprehensive Chain-of-Thought (CoT) prompting, reasoning language model, extensive tuning for tool use, often misses a large fraction of long‑tail, multi-hop facts because the optimization landscape is too stiff to plan multi‑step web navigation, branch or backtrack \cite{biran-etal-2024-hopping}, \cite{gema2025inversescalingtesttimecompute}. Agentic scaffolds that interleave reasoning and tool use, such as \textsc{ReAct} \cite{yao2023react} and \textsc{Reflexion} \cite{shinn2023reflexion}, alleviate this by decomposing the search space and learning from self‑critique. An example would be Grok‑4 Heavy, which credits their performance jump from 44\% to 50\% on Humanity’s Last Exam to parallel agentic search \cite{xai2025grok4}. We therefore compare multiple model–scaffold configurations on the end-to-end task of enumerating all competitor drugs for a given indication and show that scaffolded agents reliably outperform single-pass prompting, especially on harder, fragmented cases.

Motivated by these gaps and the scaffold vs. “bitter lesson” trade-off, we build and evaluate a competitor-discovery agent and a post-retrieval LLM-as-judge (Competitor-Validator) on a new, domain-grounded benchmark. To our knowledge, no prior public benchmark evaluates competitor discovery; accordingly, we report baselines for OpenAI Deep Research and Perplexity Labs and find that our agent achieves 83\% recall (vs. 65\% and 60\%, respectively) while the validator suppresses false positives to maximize precision. The agent is deployed as a component of a fast due diligence system used by enterprise users; in a VC case study, the analyst turnaround time fell from $\sim$2.5 days to $\sim$3 hours ($\sim$20×). 

\section{Data}
\label{sec:dataset}

Essentially, we derive three datasets from the same memos corpus:
\begin{description}[leftmargin=0em,labelsep=0.5em,itemsep=0.35\baselineskip,style=nextline]

\item[\textbf{1. Competitors Dataset}] Ground truth for enumerating competitors per indication (used to assess recall/coverage):
\[
\mathcal{D} \;=\; \bigl\{\,(\mathit{ind}_j,\;\mathcal{C}^{\star}_{\!j})\,\bigr\}_{j=1}^{M},
\qquad
\mathcal{C}^{\star}_{\!j}=\{d_{j1},\ldots,d_{jK_j}\}.
\]
Here, \(\mathcal{C}^{\star}_{\!j}\) is expert-curated from private biotech VC fund memos and is \emph{not assumed complete} and is subjective; it reflects what competitors the experts have identified in the past during their due diligence process on specific assets.

\item[\textbf{2. Attributes Dataset}] Canonical drug attributes to assess agents' ability to recover attributes for the identified competitors.
\[
\mathcal{A}
\;=\;
\bigl\{\,\bigl(d,\;\mathit{ind}^{\ast},\;\mathbf{a}^{\mathit{ind}^{\ast}}_{d}\bigr)\,\bigr\},
\qquad
\mathit{ind}^{\ast}\in \mathcal{I}\cup\{\varnothing\},
\]

where $\mathit{ind}^{\ast}=\varnothing$ denotes that no indication is provided (i.e., the attribute is indication-independent); $\mathbf{a}^{\mathit{ind}^{\ast}}_{d}$ includes: drug name aliases, modality (type), mechanism(s) of action, targets, development stage, regulatory status, therapeutic area, other indication(s), administration routes, and company information (website/description/ticker).

\item[\textbf{3. Competitor-Validator Dataset}] Pairs for tuning (prompt refinement) and applying the post-retrieval precision filter (LLM-as-judge):
\[
\mathcal{V} \;=\; \bigl\{\,(\mathit{ind}_i,\;\mathit{drug}_i,\;y_i)\,\bigr\}_{i=1}^{N},
\qquad
y_i\in\{0,1\}.
\]
Positives (\(y_i{=}1\)) are expert-confirmed competitors; negatives (\(y_i{=}0\)) are hard near-misses (e.g., umbrella terms, out-of-scope indications, withdrawn/unrelated programs). \(\mathcal{V}\) is used to tune the Competitor-Validator toward a high F1-score for both rejecting false positives and not missing relevant competing drug names.
\end{description}

\subsection{Raw Data Description}
\label{sec:data-collection}
We use due diligence memos of the private biotech VC fund (see "Original Memo" in Fig.~\ref{fig:framework-diagram} for a schematic view of memos we used). The initial corpus contained 210 unclassified memos spanning 15 years of the VC fund's work. Out of these, we selected reports created from 2019–2025 that (i) perform due diligence on companies with at least one drug asset or (ii) assess the competitive landscape of a specific indication. This filtering yielded 73 reports with 174 \emph{drug-asset – indication} pairs.

The memos vary in language and format: the authors scatter competitive-landscape details throughout free-form text and embed every kind of artifact — low-resolution plots, full-page figures, standard tables, screenshots of slide decks, and other hard-to-parse images. To handle this heterogeneity, we built a parsing agent that reads both text and images, traverses the memos’ intended hierarchical structure: company → asset → indication → competitor list—and systematically extracts competitor drugs for every mentioned indication (see "Agentic Parsing Flow" in Fig.~\ref{fig:framework-diagram}). 

Then, for each extracted drug, the Attributes Parsing Agents extract indication-dependent and indication-independent attributes.

\subsection{Hierarchical Memos Parsing}

Our multi-agent memo parser follows a hierarchical extraction workflow (see "Agentic Parsing Flow" in Fig.~\ref{fig:framework-diagram}).
First, a \textit{Drug-Assets extractor} identifies every unique drug asset analyzed in the memo.
Each asset is then passed, in parallel, to $N$ \textit{Indications extractors}, which extract all unique indications the given asset was analyzed against; each resulting indication is subsequently routed to a \textit{Competitors extractor}, which extracts the competitor drugs mentioned for the corresponding drug–indication pair.

Then an \textit{Attributes Parser} consumes each \emph{(drug, indication)} node and emits: drug name aliases, modality (type), lead indication, other indications, and administration routes, mechanism(s) of action, targets, development stage, and regulatory status, therapeutic area, and company information (website/description/ticker).

\textbf{Quality Controls \& Implementation.} The parser is based on Google’s Gemini-2.5 Pro without web browsing. We use Gemini’s caching for efficient processing across multiple extraction stages. All four extractor agents rely on template-based, JSON-schema prompts that (i) demand verbatim evidence from the memo and (ii) reject outputs lacking an explicit mention, thereby preventing hallucination while ensuring structured, fact-only extraction. For non-English memos, the pipeline involves a translator agent, based on Gemini-2.5 Pro with web browsing, to translate indications and drug names while preserving life sciences terminology.

All outputs undergo schema-on-write validation with Pydantic. While some attributes are strict, fields like \texttt{modality}, \texttt{route of administration}, and \texttt{development stage} are controlled-vocabulary (lookup) columns seeded with some possible values (e.g., for \texttt{modality}, seeds are \texttt{'small molecule'}, \texttt{'monoclonal antibody'}; for route of administration, \texttt{'oral'}, \texttt{'subcutaneous'}) but with open-world extension --- novel memo values are admitted and normalized downstream. This yields a semi-structured, schema-constrained design that mixes closed enumerations with extensible controlled vocabularies.

\subsection{Normalization and Alias Resolution}
\label{subsec:deduplication}

The parsed JSONs mirrors the original report hierarchy: \emph{asset} → \emph{indications} → \emph{competitors} → \emph{attributes} (see “JSON” in Fig.~\ref{fig:framework-diagram}). For evaluation, we drop the top-level \emph{asset} node and flatten to an indication-centric mapping: \emph{indication} → {competitor drug → attributes}; where each drug links to its attributes.

Different asset memos can refer to the same indication --- this introduced duplicates (e.g., “NHL” vs. “Non-Hodgkin lymphoma”). In order to merge different memos into one corpus, we need to ensure consistency. Therefore, we developed an alias resolution agent that sequentially checks whether a newly parsed indication matches an existing alias and merges competitor lists when appropriate. Domain experts then manually verified and corrected the resulting mappings. For drug-level deduplication, we relied on the LLM judge described in "Alias resolution via an LLM judge" at page~\pageref{textbf:alias_resolution_llm_judge}.


After cleaning and merging, we retain 105 unique indications along with a corresponding competitor drug names lists.

\subsection{Parser Benchmarking} 
We split the memos into two non-overlapping batches: 5 memos for prompt refinement and 10 unseen memos for final assessment. On the 5-memo batch we iteratively adjusted the agent configuration and prompt until the parser reproduced every author-annotated drug-asset → indication → competitor triple. We then froze the prompt and ran it once on the 10-memo batch, where it again attained 100 \% extraction accuracy; this confirmed the parser and produced the benchmark dataset used in all subsequent experiments.

\subsection{Data Splitting}
\label{subsec:data_splitting}

\begin{figure}[t]
  \centering
  \includegraphics[width=1.0\linewidth]{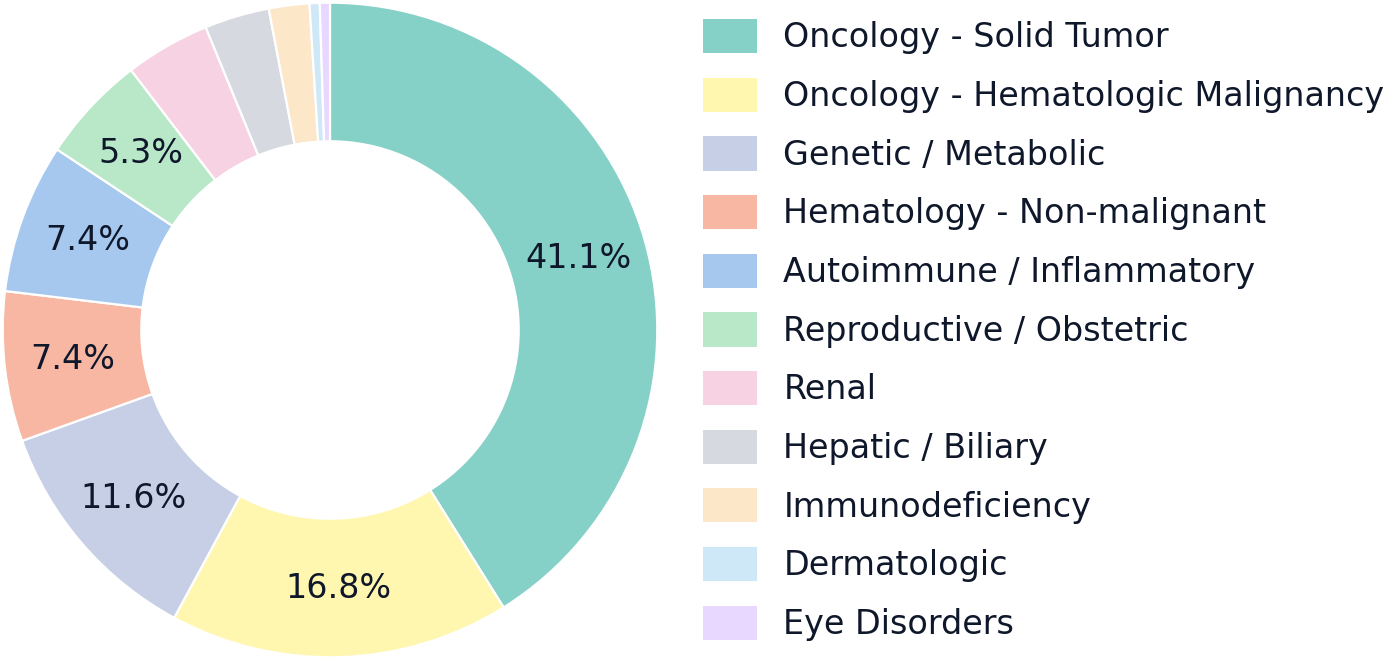}
  \caption{
    Distribution of indications clustered to general categories.
  }
  \label{fig:indication-categories}
\end{figure}

We needed a smaller test set to manually run OpenAI Deep Research and Perplexity Labs in order to mimic the behavior of our clients using the corresponding systems through the publicly available UI apps. To prioritize newer indications, we sorted all normalized entries by the latest mention date across their aliases. The 50 most recent were assigned to the test set, and the next 50 to the validation set (see "Validation/Test Split" in Fig.~\ref{fig:framework-diagram}). In Fig.~\ref{fig:indication-categories} we grouped indications into general disease groups and show the distribution over test and valid splits.

\subsection{Competitor-Validator Dataset: Negative Samples Mining}
\label{sub:precision_grader_dataset}
As mentioned in earlier sections, the resulting set of indication–competitor drug pairs can be used only to measure the recall. While this list is incomplete, it is sufficient for evaluating coverage. To compute precision, however, we must also identify false positives among the predicted competitors --- drugs that were incorrectly placed in the competitive landscape by the agent.

This requires a scalable evaluation setup that can be integrated into Continuous Integration/Continuous Delivery (CI/CD) pipelines, enabling recall and precision to be computed automatically whenever a new agent is pushed to the code base. We therefore needed a real-time grading mechanism for competitor validity. Specifically, we require a Competitor-Validator Agent that, given an indication and a drug name, could determine whether the drug belongs to the competitive landscape of that indication.

At this stage, we had only positive samples $\mathcal{C}^{\star}_{\!j}$. To iteratively refine the Competitor-Validator prompt, we also required negative samples. For each of the 50 test and 50 validation indications, we used the GPT o3-pro agent to predict competitor drug names. After removing known true competitors, we randomly sampled two remaining candidates per indication. Each resulting pair $(\mathit{ind}_i, \mathit{drug}_i)$ was passed to the original authors of the memos for expert labeling (see "Negative Sample Mining" in Fig.~\ref{fig:framework-diagram}.). The task was to determine whether $\mathit{drug}_i$ is a valid competitor for $\mathit{ind}_i$, following the criterion defined in Fig.~\ref{lst:competitor_definition}.

\begin{figure}[t]
\centering
\fbox{\parbox{0.95\columnwidth}{
\small \textbf{Definition of a Competitor Drug.}
Exclude drugs that are completely mechanistically or clinically irrelevant to the given indication, even for learning-from-failure purposes. Drugs that should be 100\% excluded include those with no relevant MoA (Mechanism of Action), no trial history, or no theoretical relevance to the disease.
}}
\caption{Definition of a Competitor Drug}
\label{lst:competitor_definition}
\end{figure}

\section{Competitor-Validator Agent}

To maximize precision for a predicted competitor list $\widehat{\mathcal{C}}_i$ for a given indication $\mathit{ind}_j$, we classify each predicted drug $d \in \widehat{\mathcal{C}}_i$ as either a true or false positive. This classification is performed by our \textit{Competitor-Validator Agent} (see "Competitor-Validator" in Fig.~\ref{fig:framework-diagram}), an LLM-as-a-judge. We use it as a filtering layer to suppress hallucinations and remove false positive drugs from the list of predicted competitors before propagating them further to the client-facing system components. This filter maintains both high recall and high precision.

\subsection{Implementation}
\label{sub:competitor_validator_implementation}

The Validation Agent was developed by iteratively refining prompts on the expert-labeled dataset described in Subsection "Competitor-Validator Dataset: Negative Samples Mining" at page ~\pageref{sub:precision_grader_dataset}. The final agent is based on the \textsc{ReAct} (Reasoning and Acting) framework with 3 attempts (\textsc{ReAct-3}) and utilizes Gemini-2.5 Flash with a web search tool.

The agent's investigation is highly structured. It is prompted to act as a specialist in pharmaceutical market analysis and is instructed to query a range of authoritative sources, such as:
\begin{itemize}
    \item Clinical trial registries (e.g., ClinicalTrials.gov)
    \item Regulatory filings and approvals (e.g., from the FDA and EMA)
    \item Scientific literature and conference abstracts
    \item Market research reports and investor presentations
    \item Company press releases on partnerships, licensing, or M\&A activity
\end{itemize}

Crucially, the agent’s final judgment is bound by a strict set of rules. It classifies a drug as a competitor only if there is verifiable evidence for the same indication: (i) clinical development or approval (including active, completed, failed/discontinued/withdrawn trials and case reports), (ii) when no clinical trial history exists, clear mechanistic relevance plus publicly documented IND-enabling and/or preclinical evidence. Conversely, it excludes drugs with only theoretical mechanistic relevance without preclinical/clinical evidence, entries with wrong/mixed-up identifiers, broad platforms without a specific candidate, and programs for related but distinct indications or different lines of therapy without a clear, direct mechanistic link.

Given an $(\mathit{indication}, \mathit{drug})$ pair, the agent follows a Thought-Action-Observation loop for \textit{three} iterations to apply these rules and consult these sources. In each cycle, it:
\begin{enumerate}
    \item \textbf{Thinks}: Analyzes previous search results and its current understanding to formulate a hypothesis or identify knowledge gaps.
    \item \textbf{Acts}: Formulates a targeted web search query based on its thought process.
    \item \textbf{Observes}: Executes the search, retrieves information, and synthesizes the findings.
\end{enumerate}

This iterative approach enables the agent to build a robust evidence base before making a final judgment. The final output is a JSON object specifying a boolean $\hat{y}_i$ label and a justification. As shown in Table~\ref{tab:validator_performance}, the agent achieves 90.4\% precision and 85.7\% recall on the test set, resulting in an F1-score of 88.0\% and demonstrating its reliability for automated precision grading.

\begin{table}[t]
\centering
\begin{tabular}{lccc}
\hline
\textbf{Split} & \textbf{Prec.\ (\%)} & \textbf{Rec.\ (\%)} & \textbf{F1\ (\%)} \\
\hline
Validation & 90.7 & 89.5 & 90.1 \\
Test       & 90.4 & 85.7 & 88.0 \\
\hline
\end{tabular}
\caption{Performance of the Competitor-Validator on validation and test splits.}
\label{tab:validator_performance}
\end{table}

\section{Agents \& Models}
We evaluated a range of models, grouped into four main categories, to benchmark performance on the competitor discovery task. All models were run with a temperature of 0.0, where supported. We used the same prompts for base models with and without web access, but employed different, specialized prompts for Deep Research agents and the LLM scaffolding frameworks to best leverage their capabilities. All systems were generally tasked to perform thorough research on competitive landscape analysis by covering sources at least mentioned in subsection "Implementation" of section "Competitor Validator" at page~\pageref{sub:competitor_validator_implementation}.

The evaluated model architectures are:
\begin{enumerate}
    \item \textbf{Models without web browsing:} Foundation models relying on their training-time knowledge, including `o3-pro` and `gemini-2.5-pro` with web browsing disabled.
    \item \textbf{Models with web browsing:} Standard API calls to foundation models with native tool use for web search, including `gpt-4o`, `gemini-2.5-pro`, and `gemini-2.5-flash`.
    \item \textbf{Deep Research agents:} Systems optimized for persistent web browsing and integrated reasoning, such as Perplexity and OpenAI Deep Research.
    \item \textbf{LLM Scaffolding Agents:} To facilitate more complex, multi-step reasoning, we implemented two agentic frameworks on top of LLM calls: \textsc{ReAct}, \textsc{ReAct} with \textsc{Reflexion}
\end{enumerate}

\textbf{\textsc{ReAct}:} To move beyond single-turn queries, we first implemented a \textsc{ReAct} agent \cite{yao2023react}. Our implementation prompts the agent to act as a "diligent and creative research agent" tasked with comprehensively mapping the competitive landscape. The agent iterates through a Thought-Action-Observation loop for a set number of attempts (N). In each loop, it analyzes its previous findings to form a search strategy, formulates a targeted web query, and synthesizes the results, allowing it to dynamically build on its intermediate findings. We denote such configurations with the suffix “\texttt{$-N$}” (e.g., \textsc{ReAct-3} uses 3 \textsc{ReAct} iterations/attempts).

\textbf{\textsc{Reflexion}:} Our experiments with \textsc{ReAct} revealed that while increasing the number of iterations improved recall, it often did so at the cost of lower precision. To address this trade-off, we introduced a meta-cognitive layer by implementing a \textsc{ReAct} with \textsc{Reflexion} framework \cite{shinn2023reflexion}. In this two-stage process, a \textsc{ReAct} "actor" agent first generates a solution. A second "reflector" agent, prompted as an expert analyst, then critiques the actor's output. The reflector is tasked with diagnosing failures, identifying incorrect competitors, and formulating a high-level plan for improvement. This structured critique is fed back to the actor to guide the next iteration, creating a loop designed to prune false positives and repeated for $M$ rounds. Where the history of actor's answers and reflections on them is propagated to the next round. We denote such configurations with the suffix “\textsc{-Reflexion-M}” (e.g., \textsc{ReAct-3-Reflexion-3} uses 3 \textsc{ReAct} iterations/attempts and 3 \textsc{Reflexion} rounds).

\textbf{\textsc{Reflexion} with \textsc{History}:} To further enhance the efficacy of this critique, we experimented with a variant, \textsc{ReAct} with \textsc{Reflexion-History}. In this setup, we provide the reflector not just with the actor's final output but also its entire reasoning trace --- the full history of thoughts, actions, and observations. We hypothesized that this deeper contextual information would allow the reflector to diagnose more subtle flaws in the actor's search strategy and provide a more effective and nuanced critique to guide the subsequent attempt. We denote such configurations with the suffix “\textsc{-History}” (e.g., \textsc{ReAct-3-Reflexion-3-History} uses 3 \textsc{ReAct} iterations/attempts and 3 \textsc{Reflexion} rounds and a History).

\textbf{Web tools:}
\emph{Observation} step of the \textsc{ReAct} accepts the query and performs the query using the web tool specified by the \emph{Act} step. We expose two retrieval tools and let the agent choose among them per step: (i) \emph{Gemini-2.5 Pro} with browsing, and (ii) \emph{Perplexity Sonar}.
To ensure that retrieved evidence is genuinely web-sourced and reliable, we pass all candidate snippets through a basic verification gate: checking whether there are groundings returned with the result, whether groundings are not pure text generations with no supports, URL resolution and recovery of the cited span (verbatim or high lexical overlap), cross-source corroboration for high-impact claims (at least two independent sources), deduplication/canonicalization of mirrors and press-wire duplicates.

\textbf{Step-parallel queries:} To increase retrieval \emph{breadth}, we enable step-parallel fan-out: at each \textsc{ReAct} step, the agent may generate up to $S$ queries \emph{in parallel}, sending each query to two search tools (Gemini-2.5 Pro, Perplexity Sonar). All returned results are merged in the \emph{Final Answer} step, which consolidates the full history across all steps. We denote such configurations with the suffix “\texttt{$-S-S_{\max}$}”; e.g., 
\textsc{ReAct-12-S-20} uses 12 \textsc{ReAct} iterations and allows up to 20 parallel web searches per step.

\textbf{Step summary.}
All “\texttt{$-S-S_{\max}$}” variants include a \emph{step-summary} mechanism. 
After each \textsc{ReAct} step, we extract a compact, structured record of facts and append it to the running state. 
For competitor discovery, this record captures the canonical drug names identified so far with their associated retrieved insights. 
These summaries normalize the otherwise heterogeneous, verbose tool outputs, reduce omission/forgetting in long histories. 
In ablations, agents with step summaries perform better than identical agents without them (see Appendix).

\section{Evaluation}
We use LLM-as-a-judge graders for both Competitors Discovery and Attributes Extraction tasks. Below, we describe these graders.

\subsection{Competitors Discovery Grader}
For each indication $i$, we compare the agent’s predicted list of competitor drug names  
$\widehat{\mathcal{C}}_i$ with an expert‐curated ground-truth list  
$\mathcal{C}_i^{\star}$.  Each ground-truth drug $d\!\in\!\mathcal{C}_i^{\star}$ is scored  
\[
r(d)=\mathbf{1}[d\in\widehat{\mathcal{C}}_i],
\]
and sample-level recall is $R_i=\frac{1}{|\mathcal{C}_i^{\star}|}\sum_{d} r(d)$.  
Mean recall over the benchmark yields the grader output  
$\mathrm{Recall}=\frac{1}{N}\sum_{i=1}^{N}R_i$.

\textbf{Alias resolution via an LLM judge:} 
\label{textbf:alias_resolution_llm_judge}
Drug names in $\widehat{\mathcal{C}}_i$ are noisy: a single API may surface as
development codes, regional brands, or salt/route strings.  
No single controlled vocabulary, whether RxNorm, DrugBank, Martindale, WHO Drug Dictionary, or EMA xEVMPD, covers the full alias tail. For example, an FDA normalization of 2024 FAERS opioid reports collapsed
7,892 heterogeneous free-text strings to just 92 RxNorm
ingredients after multiple API look-ups and manual edits --- evidence that static synonym tables buckle under real-world variability. Rule-based matching undercounts true positives in these conditions.

To avoid such alias-resolution complexity, we delegate drug alias resolution to a Gemini-2.5 Pro LLM judge with temperature set to 0. For example, for a single drug name \texttt{Utrogestan} from the ground-truth list
and a predicted set \texttt{[Atosiban, Diclofenac, ..., Nitroglycerin, Progesterone]},
the LLM judge returns the structured verdict in Listing~\ref{lst:gemini_verdict}.

\begin{listing}[t]
\caption{Structured verdict returned by the Gemini-2.5 Pro LLM judge for drug aliases resolution.}
\label{lst:gemini_verdict}
\begin{lstlisting}[language=bash]
{
  "Utrogestan": {
    "present": true,
    "matched_aliases": ["Progesterone"],
    "reason": "The drug 'Utrogestan' is a brand name for the generic drug Progesterone. The predicted list includes 'Progesterone', which is considered a match."
  }
}
\end{lstlisting}
\end{listing}

\subsection{Attributes Extraction Graders}
\label{subsec:attributes_extraction_graders}

We evaluate both indication\mbox{-}independent and indication\mbox{-}dependent attributes with the same LLM\mbox{-}as\mbox{-}a\mbox{-}judge protocol as in Competitors Discovery: all graders call \emph{Gemini-2.5 Pro} at $temperature=0$; where attribute grading requires external evidence, the judge is run with web browsing.

\textbf{Drug name aliases} grader validates each predicted alias via web-grounded checks.
The score it outputs is a precision over the validated set:
\[
\mathrm{Prec}=\frac{\lvert A_{\text{correctly predicted}}\rvert}{\lvert A_{\text{all predicted}}\rvert}.
\]

\textbf{Modality} grader checks the semantic match between the predicted modality and the reference.
The score is the binary accuracy per sample.

\textbf{Lead indication} grader tests whether the predicted lead indication semantically matches ground truth (GT). The score is the binary accuracy.

\textbf{Administration route} grader checks whether the GT route is in the predicted set, applying synonym handling. The score is the binary accuracy.

\textbf{Other indications.} For each drug, compare the ground-truth set of non-lead indications to the system’s predicted set. The per-example recall is
\[
Rec
= \frac{1}{|\mathcal{I}|}\sum_{u\in\mathcal{I}} \mathbf{1}\!\left[u\in\widehat{\mathcal{I}}\right],
\]
\noindent where $\mathcal{I}$ is the ground-truth set of “other” indications, $\widehat{\mathcal{I}}$ is the predicted set, $u$ indexes indications, and $\mathbf{1}[\cdot]$ is the indicator function. (Reported score is the mean of $R$ over the benchmark.)

\textbf{Mechanism of action} (MoA) grader decomposes the ground truth MoA into atomic statements and checks whether each statement is expressed (paraphrastically) in the prediction.
The score is the statement coverage
\[
C=\frac{1}{m}\sum_{j=1}^{m}\mathbf{1}\{\text{statement}_j\ \text{present}\}.
\]

\textbf{Target} grader normalizes names/aliases and checks if any predicted target matches any GT target. The score is the binary accuracy.

\textbf{Development stage \& regulatory status} grader compares the prediction to the most advanced historical trial state mentioned in the memo. Predictions must be $\ge$ of that state in the clinical progression lattice and status order.
The score is the binary accuracy.

\textbf{Therapeutic area} (TA) grader performs the web-grounded verification that the predicted TA matches GT for the specific drug/indication. The score is the binary accuracy.

\textbf{Company information} Company information is the composite field with \emph{three} sub-attributes: official website URL, company description, and stock ticker. Web-grounded, cascaded check: if the website is wrong, the whole attribute scores $0$; otherwise, the score is the mean of website correctness, ticker correctness, and description validity (the latter is the fraction of description statements validated online).

The score is computed in the following way:
\[
s=
\left\{
\begin{array}{l@{\qquad}l}
0, & \text{if website is incorrect,}\\[6pt]
\dfrac{s_{\text{site}}+s_{\text{ticker}}+s_{\text{desc}}}{3}, & \text{otherwise.}
\end{array}
\right.
\]

\section{Benchmarking}
\subsection{Competitors Discovery}
In this subsection, we show that specialized agents, tuned for a specific task -- in this case, competitor discovery -- outperform general-purpose AI systems. We then illustrate that although LLMs without web access can achieve high recall, removing easy samples from the test set causes recall to decline. In contrast, multi-hop LLM scaffolding with web-tool usage becomes increasingly important, showing significantly less degradation.

\subsubsection{Comparison Across Models}
\label{subsec:competitors_comparison}

In Table~\ref{tab:sys-compare}, we report only our three best-performing agents. For the full list of experiments with comprehensive analysis, refer to the subsection "Ablation Study of Competitors Discovery" in the Appendix at page~\pageref{app:ablation}. Also, here we report a production scenario, where before returning competitors' list and hence computing recall, we suppress all false positives predicted by Competitor-Validator. 

The table shows that our variant of \textsc{ReAct-12-S-20} outperforms OpenAI Deep Research and Perplexity Labs, as well as o3-pro with no web, gpt-5, gpt-4o, and gemini-2.5-pro. OpenAI Deep Research and Perplexity Labs were evaluated through their publicly available UI applications. In contrast, o3-pro, gpt-5, gpt-4o, gemini-2.5-pro were evaluated via API access; note that o3-pro does not use web search, as the API version did not support this at the time of writing.

\textsc{ReAct-12-S-20} is run with temperature 0, to ensure determinism; however, when we switch to gemini-2.5-pro's default temperature 1.0 and run 3 \textsc{ReAct-12-S-20} agents in parallel and merge their results (\textsc{ReAct-12-S-20-Ensemble-3}, which is \textsc{Bioptic Agent}), it gives the best recall of 0.83. 

\begingroup
\setlength{\textfloatsep}{8pt}
\setlength{\floatsep}{8pt}
\setlength{\intextsep}{8pt}

\begin{table}[!t]
\centering
\small
\setlength{\tabcolsep}{12pt}
\renewcommand{\arraystretch}{1.15}
\begin{tabular}{@{} l r @{}}
\hline
\textbf{System} & \textbf{Recall} \\
\hline
\textbf{Bioptic Agent} (gemini-2.5-flash) & \textbf{0.83} \\
ReAct-12-S-20 (gemini-2.5-pro) & 0.78 \\
ReAct-3-Reflexion-3-History (gemini-2.5-pro) & 0.77 \\
ReAct-3 (claude-sonnet-4) & 0.68 \\
o3-pro (no web)      & 0.67 \\
OpenAI Deep Research & 0.65 \\
gpt-5 & 0.63 \\
Perplexity Labs      & 0.60 \\
gemini-2.5-pro & 0.59 \\
gpt-4o & 0.56 \\
\hline
\end{tabular}
\caption{Competitors Discovery performance comparison across systems on test split of the Competitors Discovery Dataset. Recall is given after predicted false positives were suppressed by Competitor-Validator.}
\label{tab:sys-compare}
\end{table}
\endgroup

\subsubsection{Performance on Hard Samples}
\label{subsec:hard_samples}
To evaluate model robustness, we analyzed performance across increasingly difficult indications. We define difficulty relative to the performance of the non-web o3-pro baseline: samples on which this model performs poorly are considered more difficult. As shown in Figure~\ref{fig:sample-difficulty}, the performance of non-web models collapses on these harder samples, confirming that grounding in external knowledge is essential.

The analysis further shows that the performance gap between simple web-enabled models and the iterative \textsc{ReAct} agents widens as task difficulty increases. This suggests that for difficult indications where information is fragmented, the ability to conduct multi-hop reasoning by synthesizing evidence across multiple web search attempts is critical. In the most difficult regimes (e.g., performance of o3-pro is $\le 40\%$), a higher number of search iterations (\textsc{ReAct-5}) provides a clear advantage over fewer iterations, underscoring the value of persistence. Throughout all difficulty levels, `gemini-2.5-flash` remains competitive with or outperforms `gemini-2.5-pro`, indicating that the model size is less significant when the model is allowed to perform a few web searches.

\subsection{Attributes Extraction}

Table~\ref{tab:attributes-extraction} reports per-attribute scores using the graders (see "\emph{Attributes Extraction Graders}" subsection at page~\pageref{subsec:attributes_extraction_graders}). A lightly tuned \textsc{ReAct-12} with expert-curated prompts performs on par with OpenAI Deep Research for drug-attribute extraction, matching it on MoA and Target while edging it on Development Stage/Status, Therapeutic Area, and Company. MoA remains the most error-prone field for all systems. Performance is strong overall -- perfect on Modality and high on Administration Route -- while Lead Indication and Aliases are lower.

\begin{table}[!t]
\centering
\small
\setlength{\tabcolsep}{6pt}
\renewcommand{\arraystretch}{1.25}
\begin{tabular}{@{} l
                S[table-format=1.3]   
                S[table-format=1.2]   
                @{}}
\toprule
\textbf{Attribute} &
\multicolumn{1}{c}{\makecell{\textbf{OpenAI Deep}\\\textbf{Research}}} &
\multicolumn{1}{c}{\textbf{ReAct-12}} \\
\midrule
Overall                   & 0.76 & {\bfseries 0.82} \\ 
Aliases                   & 0.78  & 0.79             \\ 
Modality                  & 0.96  & 1.00             \\ 
Lead indication           & 0.80  & 0.76             \\ 
Administration route      & 0.90  & 0.91             \\ 
Other indications         & 0.14  & 0.43             \\ 
MoA                       & 0.61  & 0.61             \\ 
Targets                   & 0.84  & 0.84             \\ 
Status \& Stage           & 0.84  & 0.92             \\ 
Therapeutic area          & 0.92  & 1.00             \\ 
Company info              & 0.77  & 0.89             \\
\bottomrule
\end{tabular}
\caption{Attributes extraction performance (higher is better) with models as columns. See subsection ``Attributes Extraction Graders'' on page~\pageref{subsec:attributes_extraction_graders} for metric definitions and scoring rules.}
\label{tab:attributes-extraction}
\end{table}

\section{Production}

\textbf{Operational impact.}
In a private biotech VC case study, instrumenting analyst workflows for competitive scans yielded a substantial throughput gain: analyst turnaround time \emph{per drug asset} decreased from $\sim\!2.5$ days to $\sim\!3$ hours ($\approx 20\times$ faster) while maintaining analyst-verified precision. On the Competitors Discovery Benchmark (Table~\ref{tab:sys-compare}), the recall -- coverage of drug asset names $\mathcal{C}_i^{\star}$ identified by experts as in the competitive environment for the given indication -- is not 100\%; however, more importantly, the system surfaced additional relevant drug assets outside that ground truth set $\mathcal{C}_i^{\star}$. After review, VC analysts validated these previously unseen assets as correct and decision-useful, indicating improved discovery power without sacrificing precision.

\textbf{Deployment} We deploy our \textsc{ReAct} agent behind a lightweight Gradio front end to support analyst-in-the-loop review. The back end is a graph-orchestrated agent service, where nodes define the agent’s logic and edges determine execution flow.

Operationally, we enforce per-tool timeouts and budgets, per-tenant rate limits, version-controlled prompts and datasets, and structured logging/metrics. We also integrate Competitor Discovery and Attribute Extraction evaluations into our CI/CD process.

\section{Conclusion}
In this paper, we present a deployed, domain-grounded system for competitor discovery in drug asset due diligence, pairing a high-recall \textsc{ReAct}-style web agent with an LLM-as-judge “Competitor-Validator”. In a benchmark derived from real biopharma VC diligence memos, the agent outperforms general-purpose systems in recall, while postprocessing based on 'Competitor-Validator' suppresses false positives without sacrificing precision. 

Robustness analyses (see subsection "Performance on hard samples" at page~\pageref{subsec:hard_samples}; Fig.~\ref{fig:sample-difficulty}) show that non-web models and single-pass prompting agents degrade on harder indications, whereas scaffolded agents with multi-hop web search maintain performance. Because the difficulty metric proxies long-tail, under-linked competitors—the cases that drive diligence risk—this implies that the agent’s gains are largest where they matter most. Although certain single-pass agents achieve higher recall in our ablations (see Table~\ref{tab:agent_comparison} in "Appendix"), this difficulty-stratified analysis shows these wins concentrate on easy cases; their recall drops sharply on hard samples, whereas scaffolded, web-search agents maintain performance.

In production, the system substantially reduces analyst turnaround and surfaces decision-useful competitors beyond historical ground truth. 

\appendix
\section{Ablation Study of Competitors Discovery}
\label{app:ablation}
Table~\ref{tab:agent_comparison} is computed in an unfiltered evaluation regime: the Competitor-Validator is not used to suppress model outputs; instead, its signal contributes only to the precision calculation. This differs from Table~\ref{tab:sys-compare}, which reflects our deployment setting where Competitor-Validator acts as a conservative acceptance policy (low-confidence predictions are dropped). As expected, when removing a conservative filter, recall increases in Table~\ref{tab:agent_comparison}, while precision reflects Competitor-Validator’s judgment rather than pre-screening. Because the evaluation regimes (and underlying systems) differ, the two tables are not directly comparable; together they surface the precision–recall trade-off between a deployment-ready setting (filtered) and an analysis setting that measures upper-bound extraction capacity (unfiltered).

The \textsc{ReAct} with \textsc{Reflexion} configurations consistently deliver the highest F1 scores, with the top variant achieving 0.84. This demonstrates that a meta-cognitive loop for self-correction is a decisive factor in achieving robust performance. Our scaffolding framework also proves to be versatile: the \textsc{ReAct-3} + \textsc{Reflexion-3-History} agent attains the second highest recall (0.841), positioning it as a powerful "explorer" for comprehensive discovery. Other configurations, such as \textsc{ReAct-1-Reflexion-6}, \textsc{ReAct-1-Reflexion-3-S-10}, can be tuned for high precision (0.87/0.95), rivaling specialized systems. While dedicated search agents like OpenAI Deep Research can achieve the benchmark's highest precision (0.935), this specialization comes at a significant cost to recall, resulting in a lower overall F1 score than our top \textsc{Reflexion} agents. 

\begin{table}[t]
\centering
\footnotesize
\setlength{\tabcolsep}{3pt}      
\renewcommand{\arraystretch}{1.12}

\begin{tabularx}{\columnwidth}{@{}>{\raggedright\arraybackslash}X
  >{\centering\arraybackslash}p{2.6em}   
  S[table-format=1.3] S[table-format=1.3] S[table-format=1.3]@{}}
\toprule
\textbf{Model} & \textbf{Web} & \textbf{Precision} & \textbf{Recall} & \textbf{F1} \\
\midrule
ReAct-12-Reflexion-3-S-10 (gemini-2.5-pro)                 & $\checkmark$ & 0.933 & 0.840  & {\bfseries 0.884} \\
ReAct-12-S-20-Ensemble-3 (gemini-2.5-flash)                & $\checkmark$ & 0.900 & {\bfseries 0.845} & 0.872 \\
ReAct-12 (gemini-2.5-pro)                                  & $\checkmark$ & 0.865 & 0.844 & 0.854 \\
ReAct-1-Reflexion-3-S-10 (gemini-2.5-pro)                  & $\checkmark$ & {\bfseries 0.950} & 0.740 & 0.832 \\
ReAct-1-Reflexion-3 (gemini-2.5-pro)                     & $\checkmark$ & 0.864 & 0.798 & 0.830 \\
ReAct-1-Reflexion-6 (gemini-2.5-pro)                     & $\checkmark$ & 0.870 & 0.768 & 0.816 \\
ReAct-3-Reflexion-3 (gemini-2.5-pro)                     & $\checkmark$ & 0.800 & 0.815 & 0.807 \\
ReAct-3-Reflexion-3-History (gemini-2.5-pro)             & $\checkmark$ & 0.773 & 0.841 & 0.806 \\
ReAct-1 (gemini-2.5-pro)                                   & $\checkmark$ & 0.834 & 0.742 & 0.785 \\
OpenAI Deep Research                                       & $\checkmark$ & 0.933 & 0.671 & 0.781 \\
ReAct-5 (gemini-2.5-pro)                                   & $\checkmark$ & 0.757 & 0.789 & 0.773 \\
gemini-2.5-pro                                             & $\checkmark$ & 0.852 & 0.699 & 0.768 \\
o3-pro (2 candidates)                                      & $\times$     & 0.721 & 0.796 & 0.757 \\
ReAct-30-Reflexion-3 (gemini-2.5-pro)                    & $\checkmark$ & 0.695 & 0.815 & 0.750 \\
ReAct-3 (gemini-2.5-pro)                                   & $\checkmark$ & 0.727 & 0.773 & 0.749 \\
gemini-2.5-pro                                             & $\times$     & 0.814 & 0.691 & 0.747 \\
4o                                                          & $\checkmark$ & 0.910 & 0.608 & 0.729 \\
ReAct-3 (claude-sonnet-4) & $\checkmark$ & 0.730 & 0.727 & 0.728 \\
Perplexity Labs                                            & $\checkmark$ & 0.865 & 0.629 & 0.728 \\
gpt-5                                                       & $\checkmark$ & 0.835 & 0.636 & 0.722 \\
ReAct-1 (gemini-2.5-flash)                                 & $\checkmark$ & 0.629 & 0.838 & 0.719 \\
\bottomrule
\end{tabularx}

\caption{Comparison of LLM agents performance with and without web search. Agents are evaluated by Precision, Recall, and F1 on a test split of a Competitors Discovery Dataset. Results are sorted by F1. No predictions post-filtering with Competitor-Validator.}
\label{tab:agent_comparison}
\end{table}

Beyond headline scores, Table~\ref{tab:agent_comparison} shows three consistent patterns. First, adding step-parallel search and modest ensembling improves coverage without a large precision penalty: \textsc{ReAct-12-S-20-Ensemble-3} attains \mbox{R=0.845}, \mbox{P=0.900}, \mbox{F1=0.871}, indicating that breadth of retrieval—not just deeper reasoning—drives recall. Second, \textsc{Reflexion} stabilizes the precision–recall trade-off at small $N$: \textsc{ReAct-1-Reflexion-3-S-10} reaches the highest precision (0.95) at the cost of recall (0.740; F1=0.842), whereas the history variant \textsc{ReAct-3-Reflexion-3-History} maximizes recall among \textsc{Reflexion} settings (0.841) by leveraging the actor/reflector trace. Third, simply increasing iteration budgets without adding a step summary (all '$-S-S_{max}$' variants use step summary) shows diminishing or negative returns, due to context overload: \textsc{ReAct-30-Reflexion-3} raises recall (0.815) but depresses precision (0.695), lowering F1 to 0.750. General-purpose systems remain skewed—e.g., OpenAI Deep Research (P=0.933, R=0.671; F1=0.780) and \texttt{gpt-5} (P=0.835, R=0.636; F1=0.722)—underscoring the value of targeted scaffolds with verification. Practically, these results justify pairing a high-recall explorer (e.g., the history variant at small $N$ or \textsc{ReAct-12-S-20}) with the Competitor-Validator to enforce precision, and reserving high-precision \textsc{Reflexion} settings for verification passes.


\FloatBarrier
\bibliography{aaai2026}

\end{document}